\documentclass[letterpaper, 10 pt, conference]{ieeeconf}  

\IEEEoverridecommandlockouts                              

\usepackage{amsmath,amsfonts}
\usepackage{algorithmic}
\usepackage{algorithm}
\usepackage{array}
\usepackage[caption=false,font=footnotesize,labelfont=rm,textfont=rm]{subfig}
\usepackage{textcomp}
\usepackage{stfloats}
\usepackage{url}
\usepackage{verbatim}
\usepackage{graphicx}
\usepackage{cite}
\usepackage{xcolor}
\usepackage{amsfonts}
\usepackage{multirow}

\title{\LARGE \bf
Visual Temporal Fusion Based Free Space Segmentation for Autonomous Surface Vessels
}

\author{Xueyao Liang$^{1}$, Hu Xu$^{1,2}$ and Yuwei Cheng$^{*1,3}$
\thanks{$^1$ Xueyao Liang is with ORCA-Uboat, Shaanxi, 710075 China(email: liang\_xy@alumni.sjtu.edu.cn)}
\thanks{$^{1,2}$ Hu Xu is with the School of Marine Science and Technology, Northwestern Polytechnical University, Shaanxi, 710072 China, and also with ORCA-Uboat, Shaanxi, 710075 China (email: {xuhu@mail.nwpu.edu.cn})}
\thanks{
$^{1,3}$ Yuwei Cheng is with the Department of Electronic Engineering, Tsinghua University, Beijing, 100084, China, and also with ORCA-Uboat, Shaanxi, 710075 China.)
}
\thanks{* Corresponding author: Yuwei Cheng (email: {chengyw18@tsinghua.org.cn})}
}

\begin{document}

\maketitle
\thispagestyle{empty}
\pagestyle{empty}

\begin{abstract}

The use of Autonomous Surface Vessels (ASVs) is growing rapidly. For safe and efficient surface auto-driving, a reliable perception system is crucial. Such systems allow the vessels to sense their surroundings and make decisions based on the information gathered. During the perception process, free space segmentation is essential to distinguish the safe mission zone and segment the operational waterways. However, ASVs face particular challenges in free space segmentation due to nearshore reflection interference, complex water textures, and random motion vibrations caused by the water surface conditions. To deal with these challenges, we propose a visual temporal fusion based free space segmentation model to utilize the previous vision information. In addition, we also introduce a new evaluation procedure and a contour position based loss calculation function, which are more suitable for surface free space segmentation tasks. The proposed model and process are tested on a continuous video segmentation dataset and achieve both high-accuracy and robust results. The dataset is also made available along with this paper.

\end{abstract}

\section{INTRODUCTION}

ASVs have gained popularity in recent years due to their potential applications. These vessels can operate in various environments, ideal for oceanographic research, environmental monitoring, and search and rescue operations \cite{zhuang2021navigating}. A high-quality perception process is crucial for the success of ASVs missions. When facing congestive ground traffic, the ASVs could be a critical alternative to the nearshore transportation system \cite{wang2023roboat}. 
However, the inland waterways, which are more complex and narrow than the open waters, pose more challenges to the ASVs perception process. 

The free space segmentation tasks are about figuring out the safe and available operation space for Autonomous Vehicles (AVs), enabling the AVs to understand the surrounding environment better and carry out efficient autonomous tasks \cite{sless2019road}. The free space segmentation on the road, which is of vital importance for Autonomous Ground Vehicles (AGVs), has attracted extensive attention \cite{Fan_2022}, \cite{2018Segmentation}. As for ASVs, the free space segmentation tasks mainly aim to distinguish water surface and shorelines. Visual perception modules are widely applied to achieve high-quality free space segmentation for ASVs and have been shown to be more cost-effective and information-dense than the sensor-based approaches \cite{kristan2015fast}. High-quality visual segmentation models have good robustness in visual hashing scenes and improves semantic segmentation at the water surface edge, which is significant for the perception system of autonomous vessels. 

\begin{figure}[!tbp]
\centering
\includegraphics[width=3.5in]{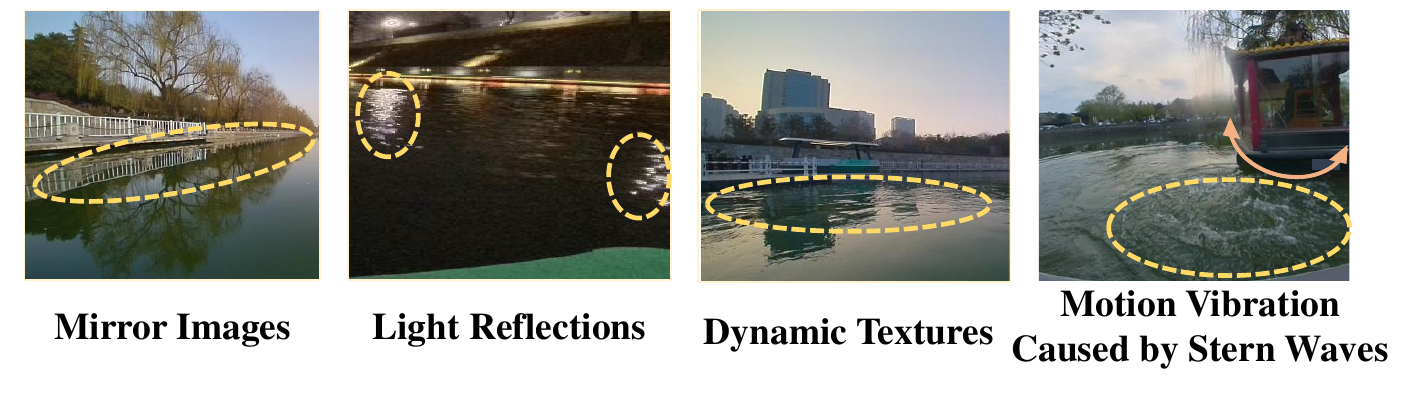}
\caption{The challenges of nearshore reflection areas remains to be solved. Existing models can not cope well with intense light, mirror images, ASV motion vibrations, and dynamic water textures, while these issues are essential to ASVs' free space segmentation tasks.}
\label{fig_reflectioncom}
\vspace{-0.2cm}
\end{figure}

However, even though there are similar task demands, Unlike ground lanes, which is primarily black and easy to separate from background objects, water surfaces exhibit varying distinctions with environmental changes, such as water level changing, and surface floating such as debris. The changes will lead to risky collisions, especially in narrow lanes. On the other hand, most ASVs require long control responding time due to underactuated kinodynamic systems with large inertia \cite{8281087}. Given that, it is essential to ensure sufficient time for control response, even when the vessel has long distance to obstacles. Therefore, high-accuracy free space segmentation plays an essential role to collision-free path planning and prediction \cite{zhang2021research}. Thus, applying visual free space segmentation for ASVs on the water surface scene faces new challenges, as illustrated in Fig. \ref{fig_reflectioncom}:
\begin{itemize}
\item[$\bullet$] Firstly, the reflection interference. Autonomous surface tasks encounter the mirror image of waterways, unlike ground scenarios. Identifying mirror images from real ones has always posed a challenge for semantic segmentation tasks. It is difficult to distinguish the illusory and real scenes.
\item[$\bullet$] Secondly, dynamic water surface textures. Complex water surface appearance can introduce intense light interference, further complicating the segmentation tasks. 
\item[$\bullet$] Thirdly, the motion vibrations. Unlike other vehicles, the ASVs face motion vibrations due to the unpredictable status of the water surface, which will bring severe interference to the perception process.
\end{itemize}

To improve the performance of free space segmentation tasks for ASVs against the challenges, some researchers studied image-based free space segmentation models in public segmentation datasets\cite{zhan2017effective}, \cite{zou2020novel}. Although the image-based models generate good performance in static conditions, these models can not consistently achieve robust free space segmentation for ASVs under scenes with complex interference. Considering the continuous spatial association for the reflection interference, mirror images, and dynamic characteristics of waterways, we note that the temporal fusion free space segmentation model may potentially reduce the various noises and improve the robustness of safe space segmentation on the water surface. Instead of taking images as input, the temporal fusion segmentation model takes image sequences as input and will fully utilize the visual information gathered. During the temporal fusion process, the fundamental problem is effectively combining the previous and current image frames. Applying richer visual information from multiple frames can eliminate interference and obtain high-precision free space segmentation results for ASVs. However, the ASVs face irregular severe vibrations due to the water surface's unpredictable status, significantly when affected by stern waves of other surface vessels. Additionally, the motion vibrations of ASVs are different from the aerial vehicles, while the amplitude of surface vehicles is much more severe \cite{9779466}, which therefore brings challenges to applying the temporal fusion segmentation model in the water surface free space segmentation tasks. 
\begin{figure}[!htbp]
\centering
\includegraphics[width=3.5in]{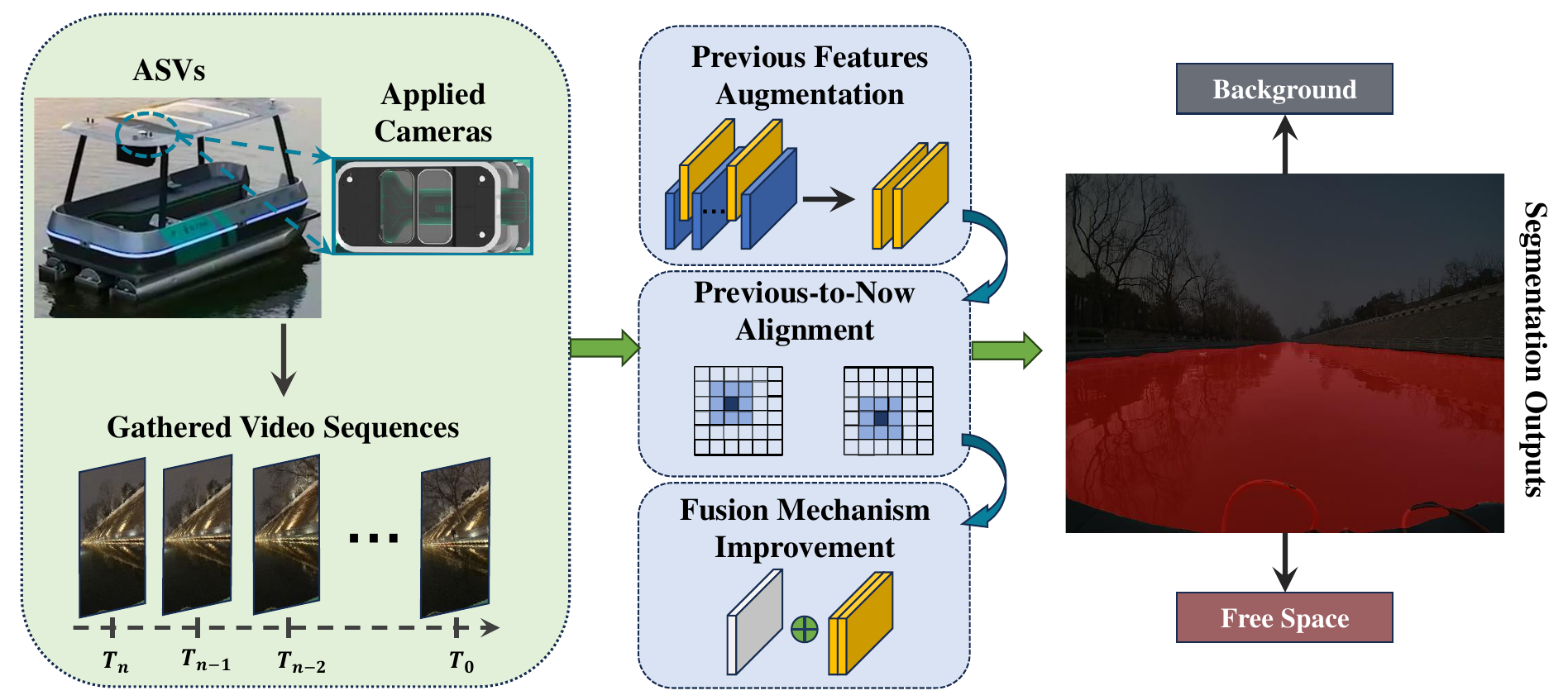}
\caption{The overall framework of this work. Our model aims to solve the environments and motion challenges during ASVs tasks and output high-quality segmentation results of free space segmentation.}
\label{fig_introframe}
\vspace{-0.2cm}
\end{figure}

In this work, to overcome the challenges in the water surface free space segmentation tasks, we propose a new visual temporal fusion based model that fulfills the gaps of ASVs' free space segmentation tasks, with high-quality performance on challenging conditions like reflection interference, dynamic water textures, and motion vibrations. Different from the existing image-based ASVs segmentation models\cite{kristan2015fast}, \cite{yao2021shorelinenet}, \cite{9755264}, we first utilize the temporal fusion model, which provides a pre-fusion augmentation module as well as a feature fusion mechanism to integrate different temporal frames properly. In addition, we designed a new loss function, which is better suited for segmentation tasks with continuous and constant contours like shorelines. Additionally, given the lack of a video sequence based dataset for ASVs free space segmentation tasks, to better evaluate the performance of our works, we built and released a video sequence based ASVs free space segmentation dataset, including 10 video sequences that consist of 5530 frames. The dataset was captured under different weather and time conditions. We also introduce a more suitable strategy for ASVs free space segmentation evaluation and test the proposed model on real-world experiments.

In conclusion, our contributions can be described as follows:
\begin{itemize}
\item[$\bullet$] We apply the visual temporal fusion mechanism in ASVs free space segmentation, proposing a new model that combines previous-to-now alignment and spatial feature augmentation procedures; 
\item[$\bullet$] Based on the characteristic of ASVs free space segmentation, we introduce a specially designed loss function and evaluation procedure, which can better serve the ASVs free space segmentation tasks;
\item[$\bullet$] The proposed model is tested on scene video datasets and achieves better performance than other images and video semantic segmentation baselines. In addition, the new video sequence based dataset is released along with this paper to benefit the ASVs research community.
\end{itemize}

\section{Related Works}

\subsection{ASVs visual Perception}
The visual perception approaches provide high-level semantic information, which enables the ASVs' safe navigation in complex surrounding environments. With the continuous advancement of image processing technology, the visual perception tasks for ASVs include target detection\cite{7382552}, target tracking\cite{wang2015vision}, and free space segmentation\cite{liu2016fast}. While the target detection and target tracking mission provide object-level semantic information, the free space segmentation tasks will generate more accurate pixel-level segmentation results for ASVs in proximity operation scenes, such as harbors and narrow inland rivers. In general, the success of the free space segmentation procedure plays a fundamental role in enabling {efficient path planning and effective decision-making during the mission \cite{zhang2021research}.

\subsection{Free Space Segmentation for ASVs}

As for AGVs, the free space segmentation on the road is extensively studied. The previous works include nearby vehicle detection \cite{wang2022review}, road characteristics and texture study\cite{qin2021light}, lane edge study\cite{qin2020ultra}, pedestrian detection\cite{leibe2005pedestrian}, etc. Although it has been extensively studied for AGVs, the free space segmentation for ASVs encounters more challenging problems on water surfaces for the complex surface characteristics. Many existing works try to improve the segmentation accuracy by focusing on sky-line distinction \cite{huiying2015water},  water texture identification \cite{wei2016effective}, cluttered background recognition \cite{zhan2017effective}, and varying environments adaptation \cite{zou2020novel}. Besides, some segmentation approaches try to learn richer, useful image information \cite{bovcon2018stereo} and increase the number of inputs \cite{bovcon2019mastr1325} for better segmentation performance. For example, Kristan et al. \cite {7073635} propose a Markov random field framework to achieve free space segmentation for diverse scenarios and regions. Shi et al.\cite{shi2019obstacles} introduce a network to extract obstacle features and enhance details in environments with cluttered backgrounds to find safe operation paths using satellite images. And Yao et al. \cite{yao2021shorelinenet} propose a deep learning-based network named ShorelineNet that applies a symmetrical encoder and decoder, claiming that the network proposed can cope with dynamic environments with high real-time performance. 

However, to our knowledge, most works applied on real-world water surfaces failed to thoroughly utilize the continuation of input information. They thus can only partially satisfy the real-world ASVs application needs. To achieve robust, safe segmentation performance for ASVs, the application of continuous vision information also remains to be explored.

\subsection{Video Semantic Segmentation and Temporal Fusion}
Video segmentation, which aims to classify pixels into different categories based on semantic information\cite{wang2021survey}, has been studied with the purpose of accuracy \cite{nilsson2018semantic} and faster calculation \cite{hu2020temporally}. The temporal fusion segmentation mechanism, which proved to be both practical and cost-saving in improving the accuracy of video segmentation, have been utilized in normal scenes \cite{sun2019multi}, \cite{yin2020fusionlane} and proven efficient for complex ground environments. In temporal fusion segmentation, the core work is to fuse the previous and current frames with numerous effective alignment and sectional spatial imaging differences \cite{qin2022uniformer}. As the video inputs present apparent irregular motion, the temporal fusion mechanism will make it challenging to integrate the temporal characteristics \cite{li2022bevformer}. On the water surface, the vibrating camera will bring challenges for the temporal fusion semantic segmentation for ASVs, where the dynamic temporal association approaches will be worth digging into to improve the temporal fusion segmentation performance in moving robotics scenes.

\section{Methodology}
To utilize continuous visual information, augment desired features, and mitigate interference, we propose a new temporal fusion based mechanism, which includes previous-to-now feature alignment and improved fusion mechanism. Additionally, to focus the model on shoreline areas, we designed a contour position based loss calculation strategy that consists of the Cross-Entropy Loss, Dice Loss \cite{li2019dice}, and 
 a new loss function that applies the contour position information; we named it Contour Loss. The whole model is as Fig. \ref{fig_2} presented.
\begin{figure*}[htbp]
\centering
\includegraphics[width=7.25in]{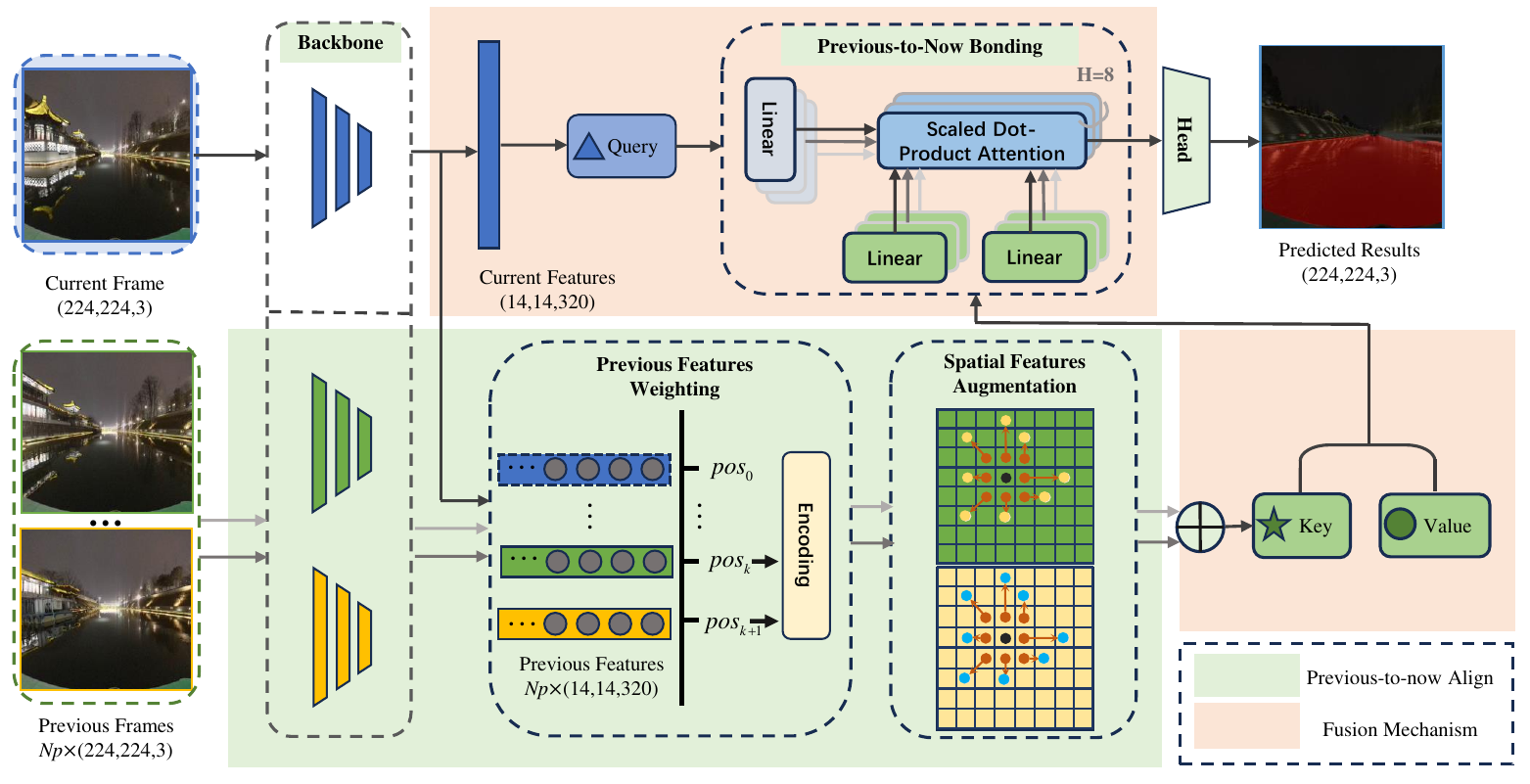}
\caption{The overall model presented by this work.} 
\label{fig_2}
\vspace{-0.45cm}
\end{figure*}
\subsection{Previous-to-now Relationship Alignment}
Our model applies MobileNet2 \cite{8578572} as the backbone. With the input video sequence of size $(224,224,3)$, the features of each frame are extracted into features with size of $(14,14,320)$. At this point, it's essential to make some adjustments to the features of previous frames by weighting, augmenting, or weakening them before going ahead with the fusion procedure to mitigate the unwanted noises. Since the majority of the data is collected while the ASVs are in motion, it will be highly probable that the input videos contain random noise, especially near the edges where close to the shorelines. This noise can lead to segmentation errors. It's also worth noting that different previous frames may have different temporal relationships with the current frame, the different temporal positions of previous frames should be indicated to describe the relevance between previous frames and the current frame.

To align each previous frame's features for more desirable features, weighted parameters need to be set to describe the temporal interval between the previous and current frames. Inspired by the position encoder\cite{vaswani2017attention}, we present a frame time interval based position encoder calculation strategy.
Through the time position encoder module, all the extracted information from the previous frame is involved in the pre-fusion augment process and weighted based on the temporal position. Next, deformable convolution \cite{dai2017deformable} is applied to the previous frames' to deal with the motion vibrations features. Eq. \ref{dcnEQ} describes this process:
\begin{equation}
\begin{split}
\label{dcnEQ}
\mathbf{F_{pre}}=\sum_{j=i}^{i+N_p}\mathrm{DCN}(\mathbf{Y_j})\cdot PE_j,
\end{split}
\end{equation}
where $\mathrm{DCN(}\mathbf{Y_j})$ denotes the frame $\mathbf{Y_j}$ with deformable convolution procedure, $PE_j$ denotes the frame time interval based position encoder, and $N_p$ denotes the number of previous frames to be fused. The overall pre-fusion approaches dealing with previous frames are illustrated in Fig. \ref{fig_prefusion}. The features are, through this procedure, augmented or weakened.
\begin{figure}[!htbp]
\centering
\includegraphics[width=3in]{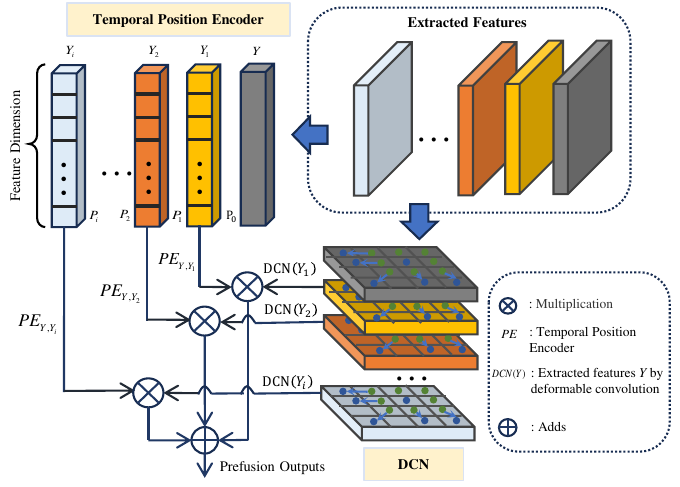}
\caption{The previous-to-now alignment module, including the temporal position encoder and deformable convolution. The selected previous frame features are, therefore, extracted and augmented or weakened.}
\label{fig_prefusion}
\vspace{-0.5cm}
\end{figure}
\subsection{Fusion Mechanism}
With temporal position encoder and deformable convolution, the temporal relevance to the current frame and the noises of the former frame itself are deliberated. In the upcoming phase, we'll apply a fusion mechanism that relies on multiple attention modules. This mechanism will amplify the correlation between the current frame features $\mathbf{F_x}$ and the previous features $\mathbf{F_{pre}}$. Additionally, the spatial and temporal features will be enhanced during the fusing process.

Inspired by multi-head attention \cite{vaswani2017attention}, we propose a multi-head cross attention that separately calculates the weighted matrix by the extracted current frames features and selected previous frames features after pre-fusion augmentations. Multi-head cross attention works by dividing the input data into multiple heads, each focusing on a different aspect of the feature data. These heads then interact through cross-attention mechanisms, allowing the features to share information and learn from each other. In our work, The multi-head cross attention module takes the information from
$\mathbf{F_{pre}}$ and $\mathbf{F_x}$ as input, the query $\mathbf{Q}$, key $\mathbf{K}$, and value $\mathbf{V}$ of the module are calculated separately by previous frames' features $\mathbf{F_{pre}}$ and currents ones $\mathbf{F_x}$. Therefore, the features extracted from both the present and previous frames are implicitly related. To enhance the desired features, a spatial attention head is utilized on the data fusion output produced by multi-head cross attention.

The fusion and augmentation modules could improve the imperfect segmentation results of nearshore mirror reflections, motion vibrations and dynamic water textures. Besides the model design, we also present a new contour position based loss calculation strategy to emphasize the prediction outputs in shoreline areas through contour position analysis. Such loss calculation is more suited for the ASVs' free space segmentation tasks.
\subsection{Loss Function}
An innovative module for calculating loss is designed based on the distances between the ground truth and output shoreline contours. It is easy to see that the performance of the free space segmentation model corresponds to the accuracy of the shoreline contour. Therefore, we introduce a contour distance and position based Contour Loss function, denoted as $L_{con}$, which calculates the fitting of predicted results and ground truth, as illustrated in Fig. \ref{fig_dl}. Minimizing the average distance can improve the gap between the predicted contour and ground truth.
\begin{figure}[htbp]
\centering
\includegraphics[width=2.5in]{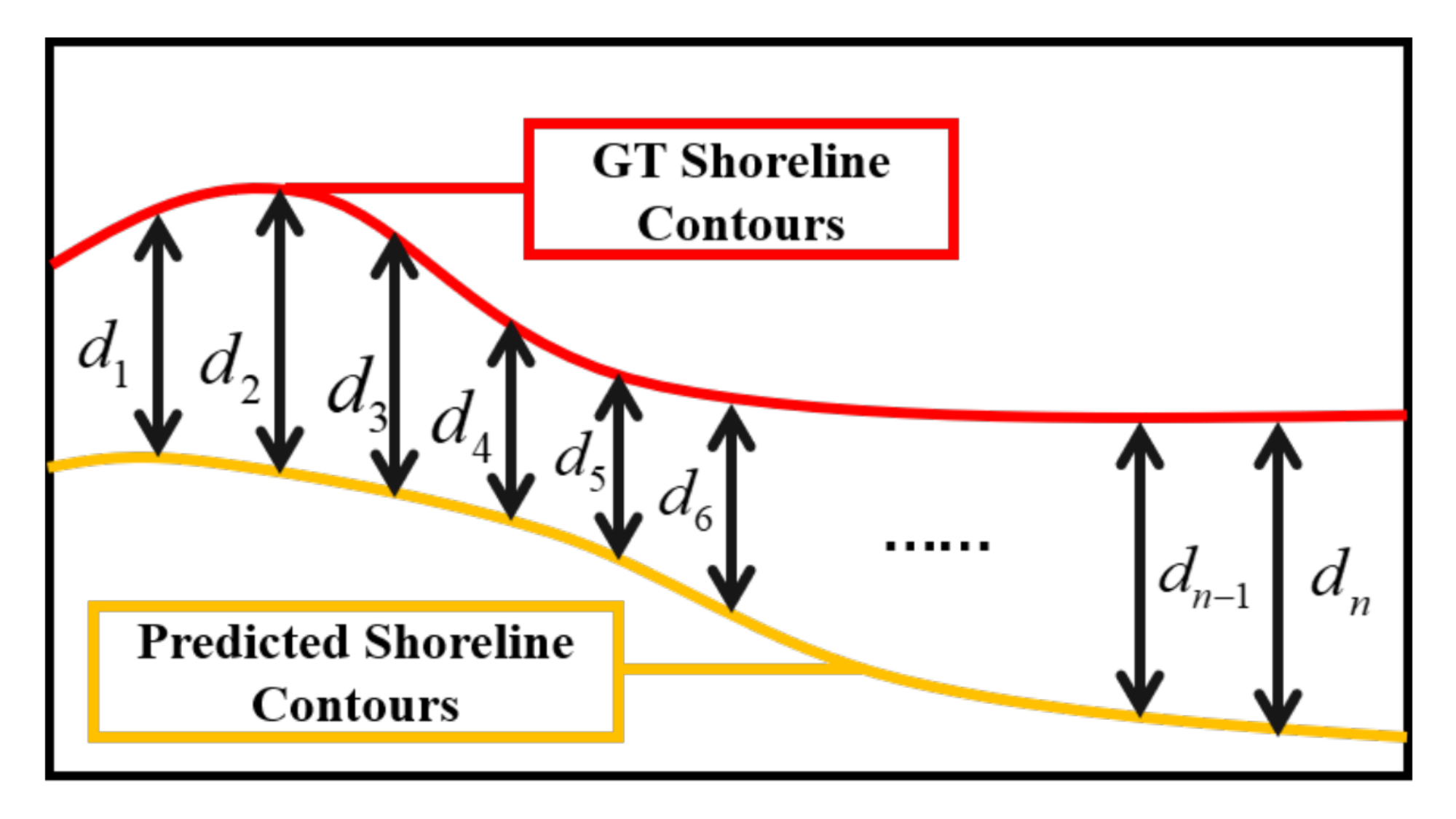}
\caption{The illustration of the contour distance between ground truth and model prediction.}
\label{fig_dl}
\vspace{-0.1cm}
\end{figure}

The distance between the predicted and ground truth will be collected through uniform sampling. The average of the sampled distance will then be generated, with a coefficient $\beta$ multiplied, to be in the same order of magnitude as the Cross-Entropy Loss ($L_{CE}$) and Dice Loss ($L_{dice}$) of the model. The Dice Loss is designed based on the dice coefficient and is calculated by taking the ratio of the intersection of the two sets to the total number of elements in the sets. These three losses will add up to the overall loss function, as Eq. \ref{disL}, where $n_c$ denotes the number of samplings, and $d_i$ denotes the sampled distance between outputs contours and ground truth:
\begin{equation}
\begin{split}
\label{disL}
L_{con}&=\frac{\beta}{n_c}\sum_{i=1}^nd_i, \\
Loss&=L_{CE}+L_{dice}+L_{con}.
\end{split}
\end{equation}

By adding up the $L_{CE}$, $L_{dice}$, and our Contour Loss $L_{con}$, the shoreline's geometrical characteristics are involved in consideration. And the training process is also made more efficient this way.
\section{Experiments and Results}
\subsection{Dataset}
After reviewing existing ASVs free space segmentation datasets in the early stages, we found that existing datasets, including our previously released work \cite{cheng2021we}, \cite{Cheng_2021_ICCV}, mostly focus on single images scenes, which can not fully satisfy the needs of video-based continuous ASVs free space segmentation tasks. As a result, in order to help with the lack of video sequence based ASVs operation datasets and evaluate our proposed model,  we hereby introduce a video sequence dataset for ASVs free space segmentation. 
\begin{figure}[htbp]
  \vspace{-0.35cm}
\centering
\subfloat[]{  
  \includegraphics[width=0.35\linewidth]{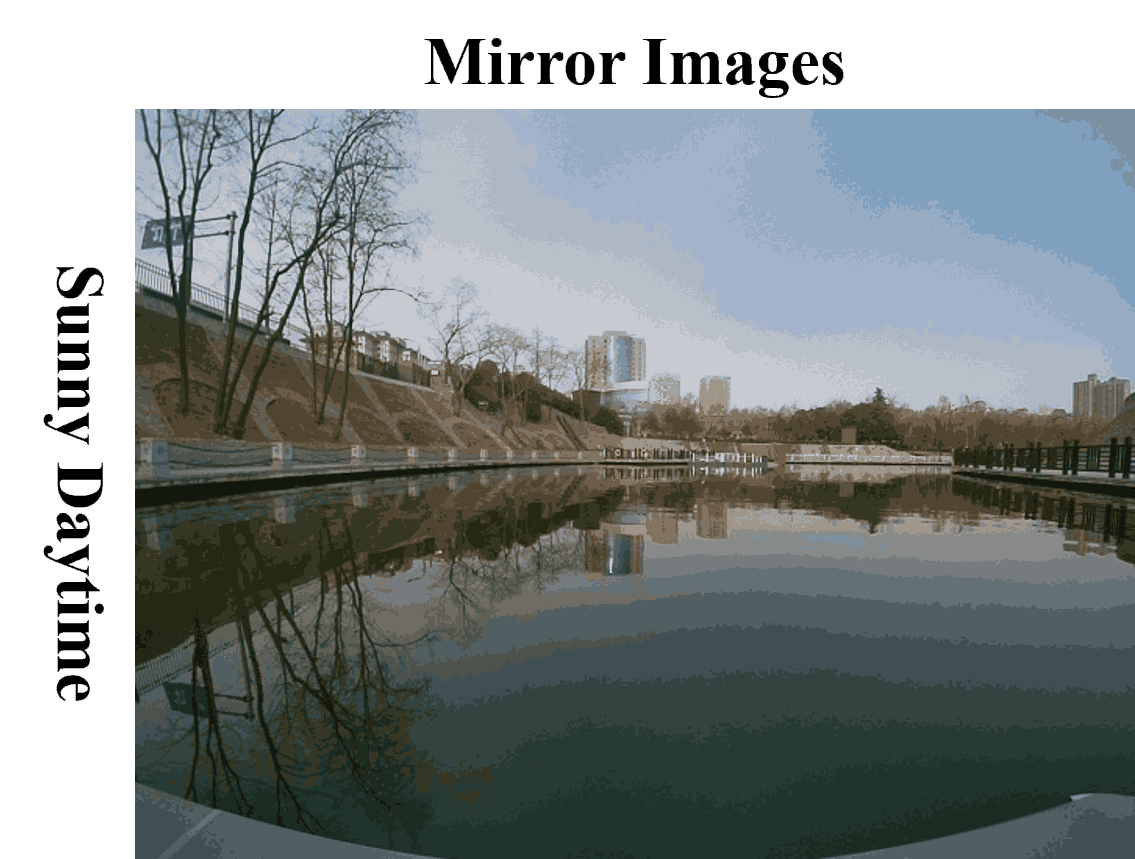}}
\subfloat[]{
  \includegraphics[width=0.305\linewidth]{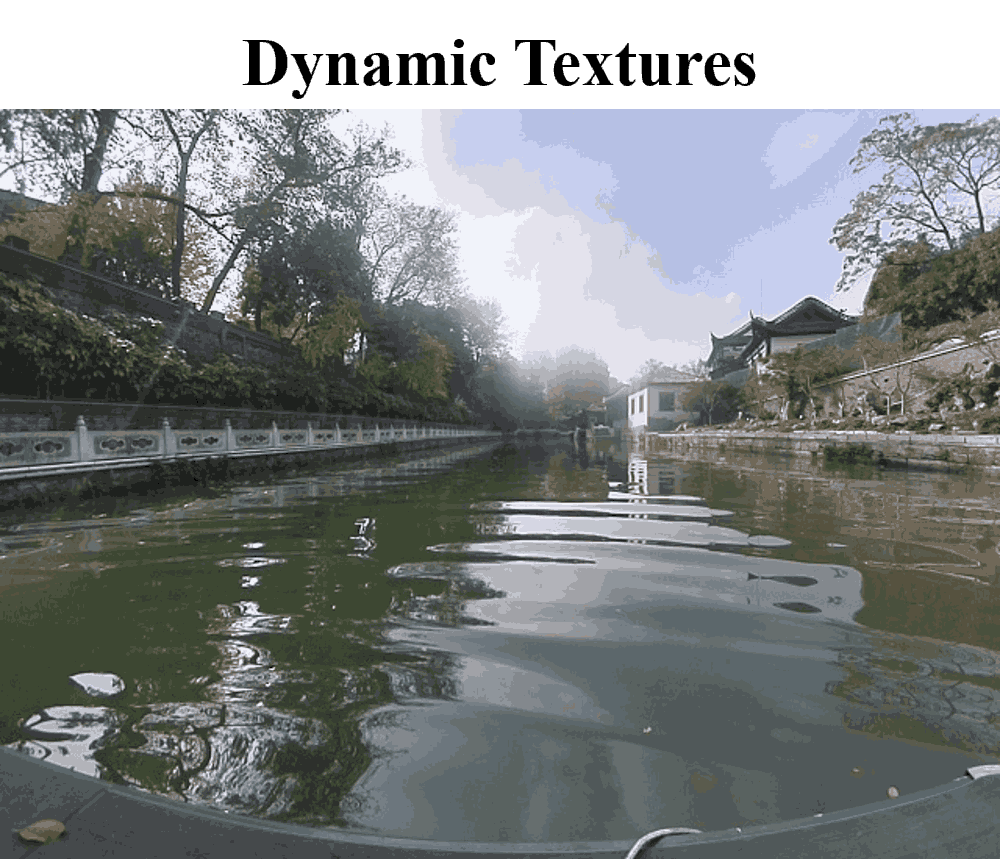}}
\subfloat[]{
  \includegraphics[width=0.305\linewidth]{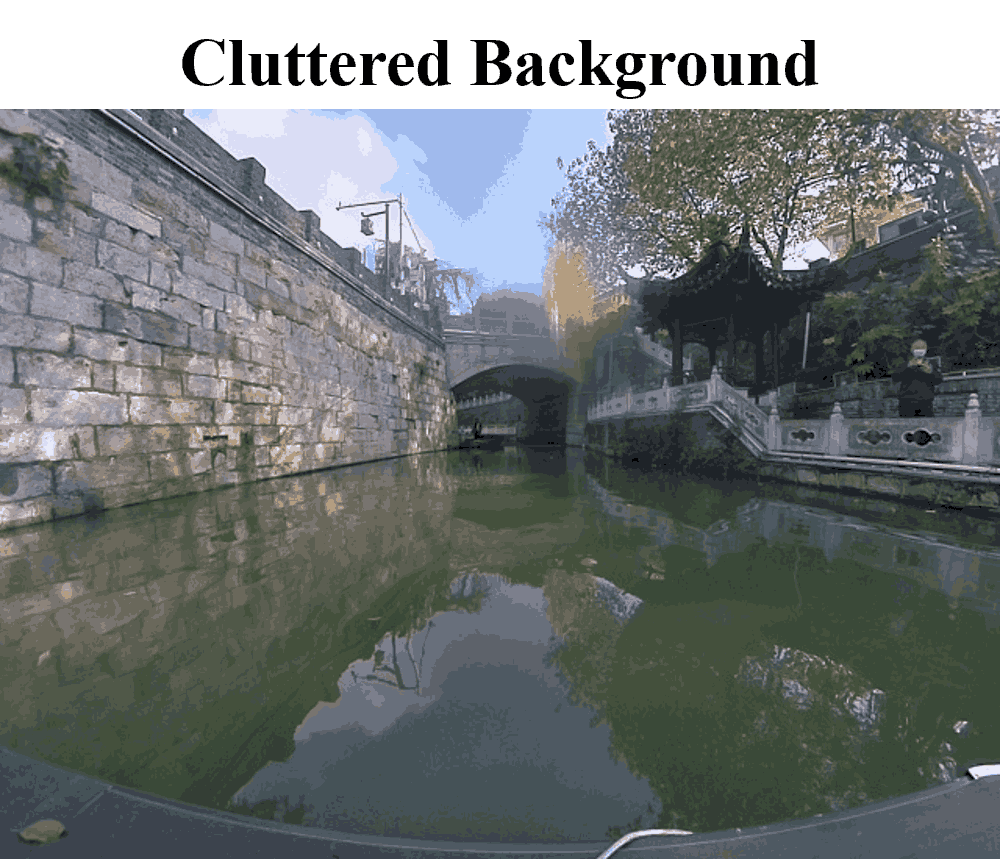}}
  \hfill
  \vspace{-0.35cm}
\subfloat[]{
  \includegraphics[width=0.35\linewidth]{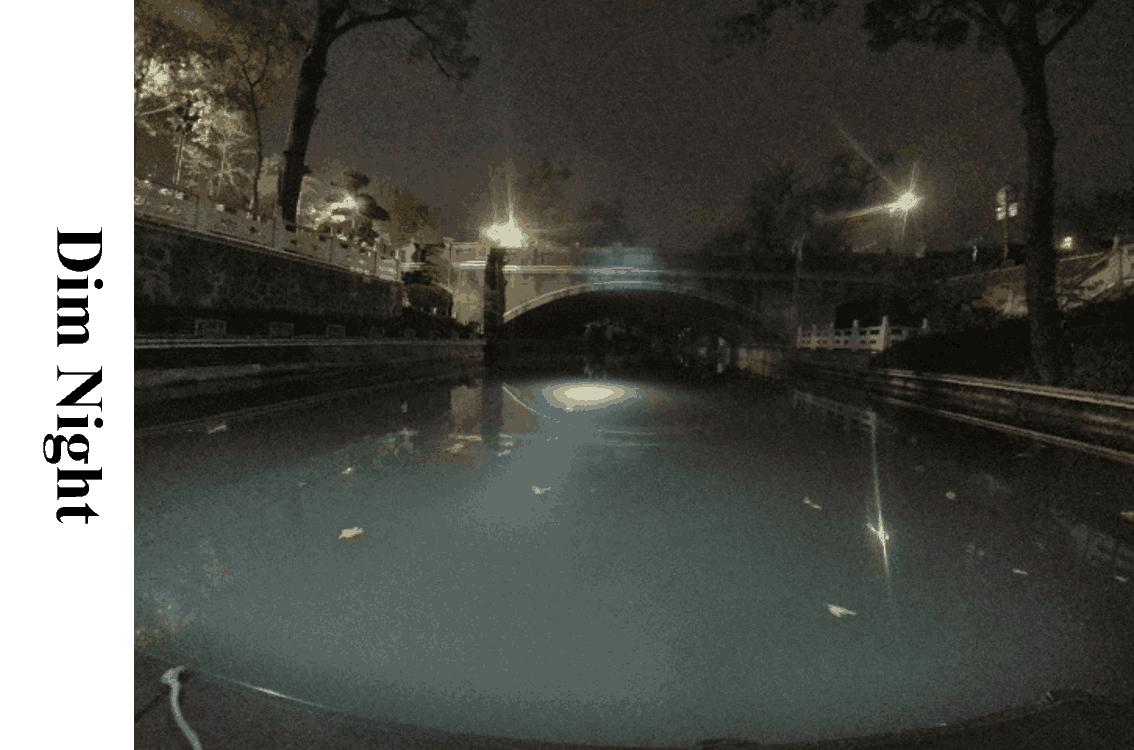}}
\subfloat[]{  
  \includegraphics[width=0.305\linewidth]{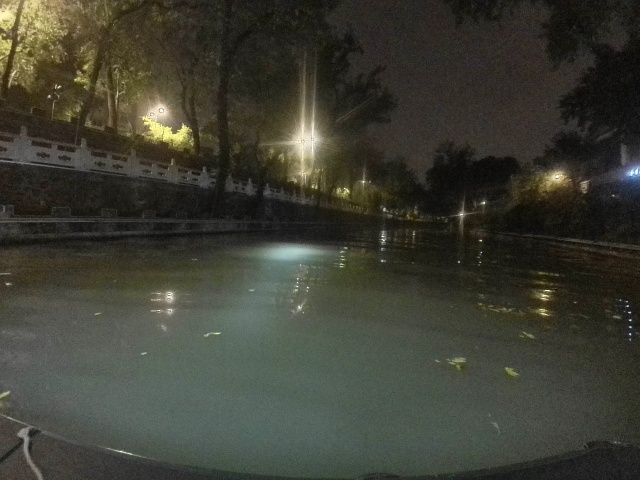}}
\subfloat[]{
  \includegraphics[width=0.305\linewidth]{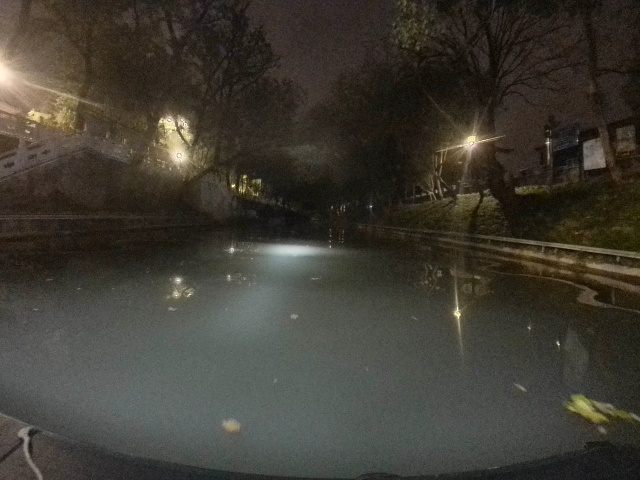}}
  \hfill
  \vspace{-0.35cm}
\subfloat[]{
  \includegraphics[width=0.35\linewidth]{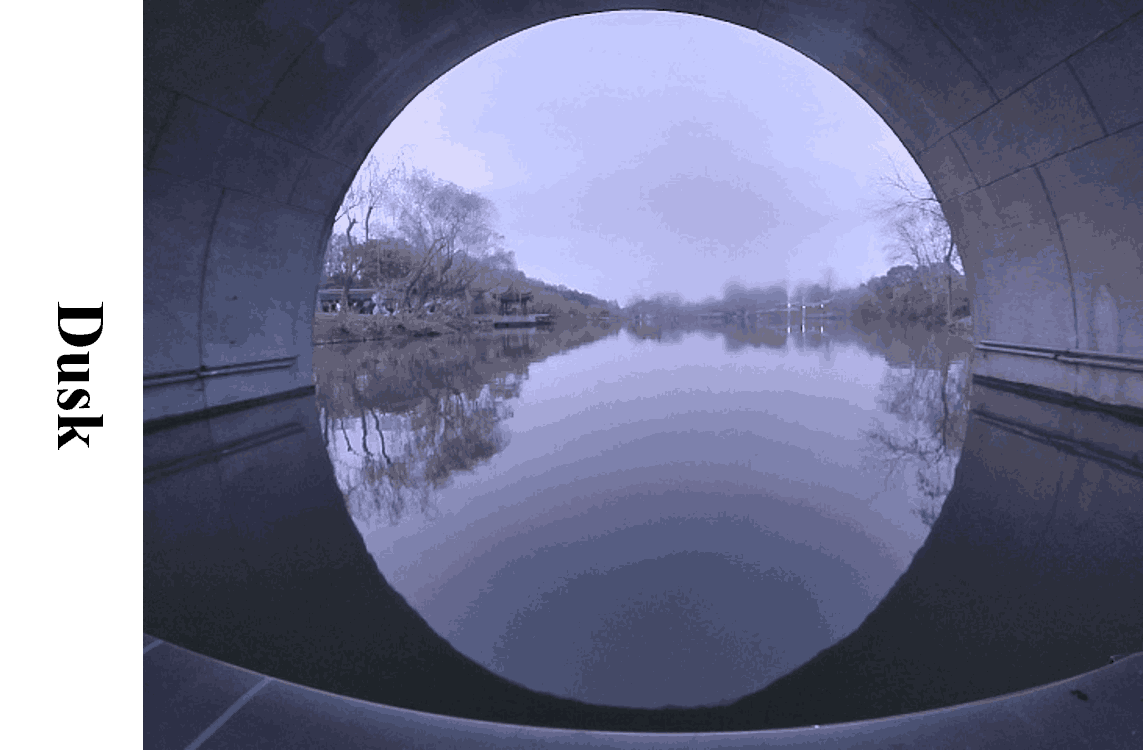}}
\subfloat[]{
  \includegraphics[width=0.305\linewidth]{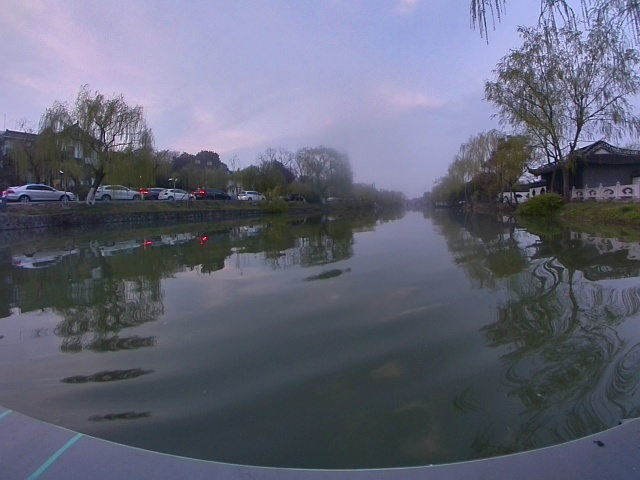}}
\subfloat[]{  
  \includegraphics[width=0.305\linewidth]{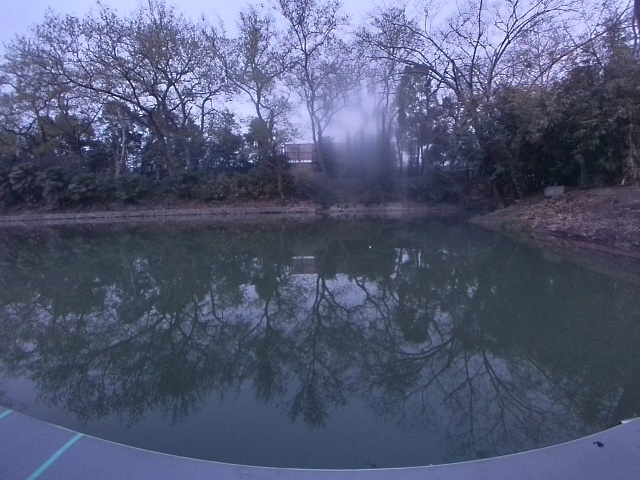}}
\vspace{-0.35cm}
\hfill
\subfloat[]{
  \includegraphics[width=0.35\linewidth]{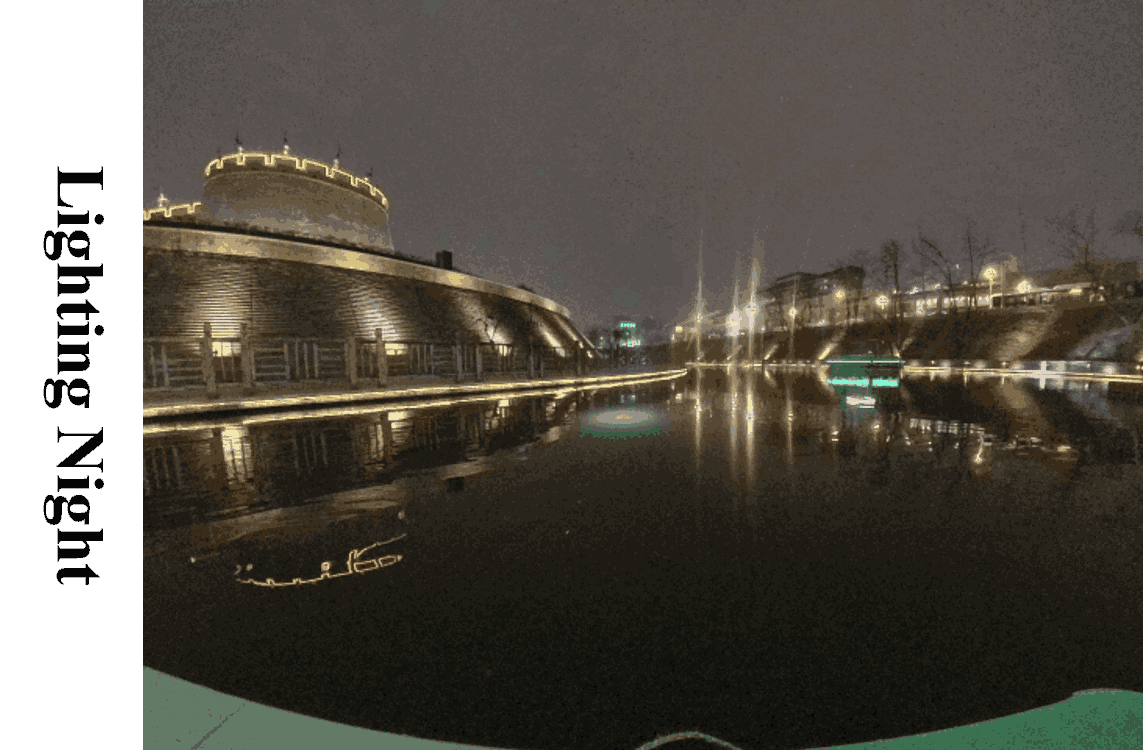}}
\subfloat[]{
  \includegraphics[width=0.305\linewidth]{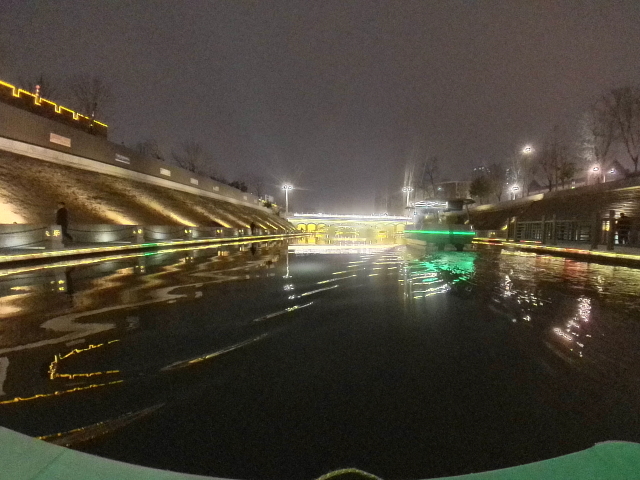}}
\subfloat[]{
  \includegraphics[width=0.305\linewidth]{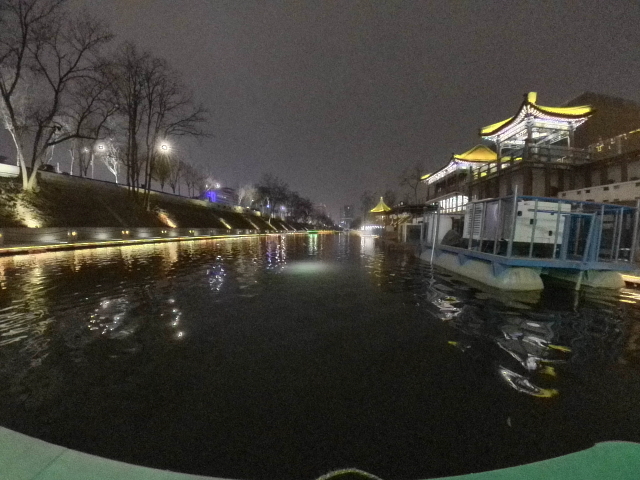}}
\caption{In the example of our introduced dataset, environments varying include sunny daytime(a-c), dim night(d-f), dust(g-i), and intense light night(j-l). To be specific, (a)(d)(g)(j) illustrate the mirror image in a sunny day, dim night, dusk, and instance lighting scene; (b)(e)(h)(k) illustrate the dynamic textures of surfaces; (c)(f)(i)(l) illustrate cluttered backgrounds.}
\label{datasetex}
\vspace{-0.1cm}
\end{figure}

\begin{figure}[htbp]
  \vspace{-0.35cm}
\centering
\subfloat[]{
  \includegraphics[width=0.32\linewidth]{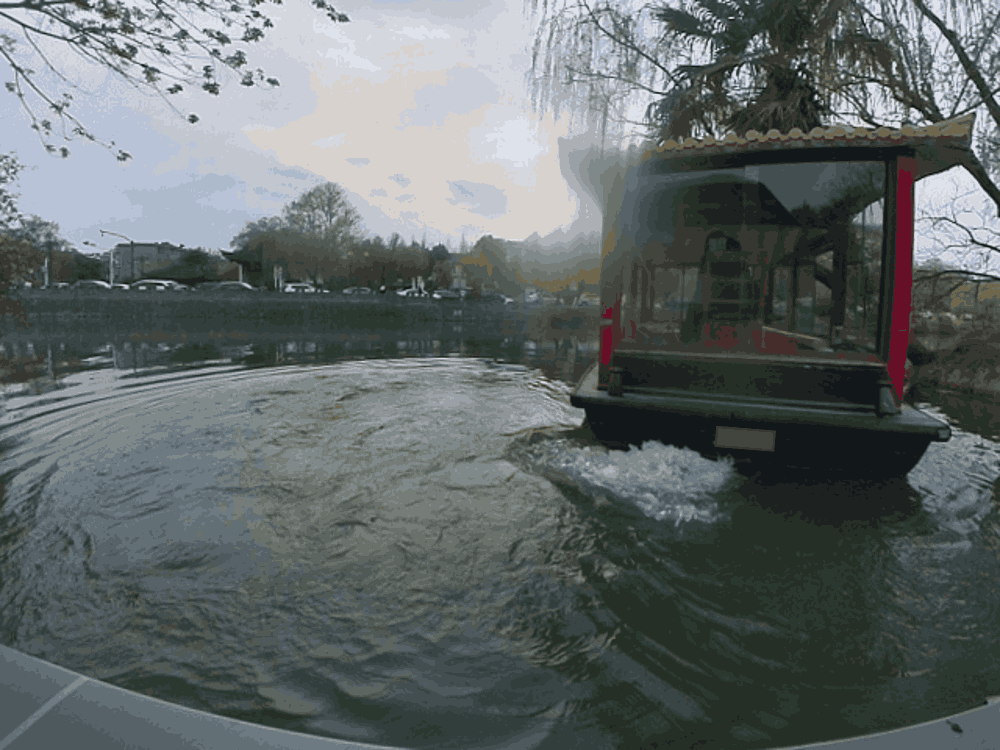}}
\subfloat[]{
  \includegraphics[width=0.32\linewidth]{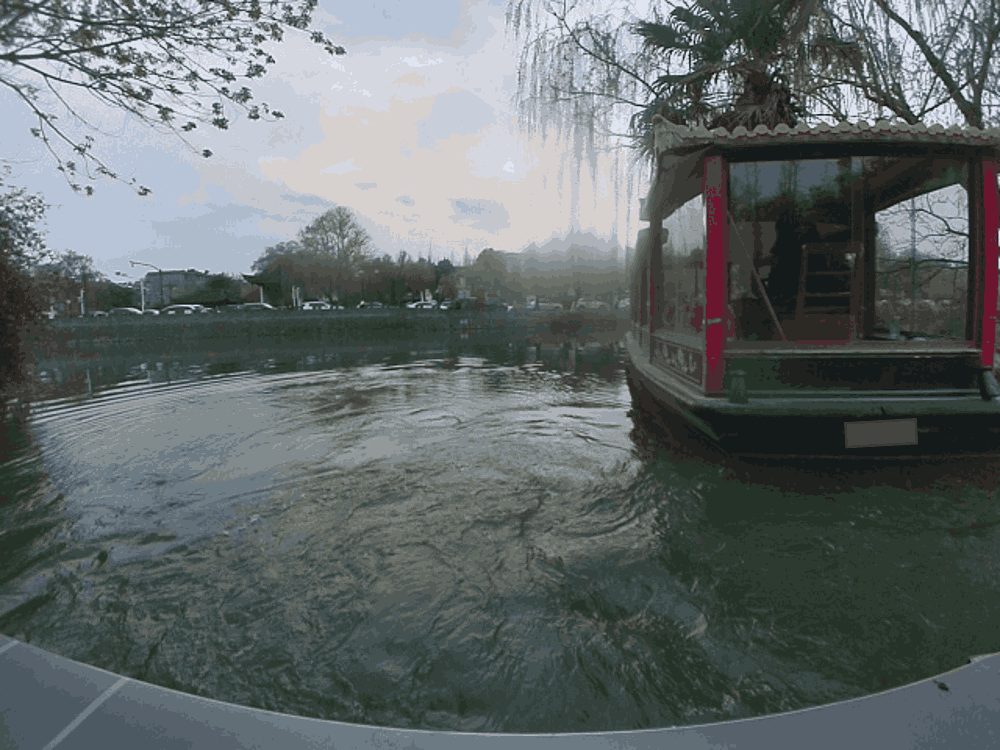}}
\subfloat[]{
  \includegraphics[width=0.32\linewidth]{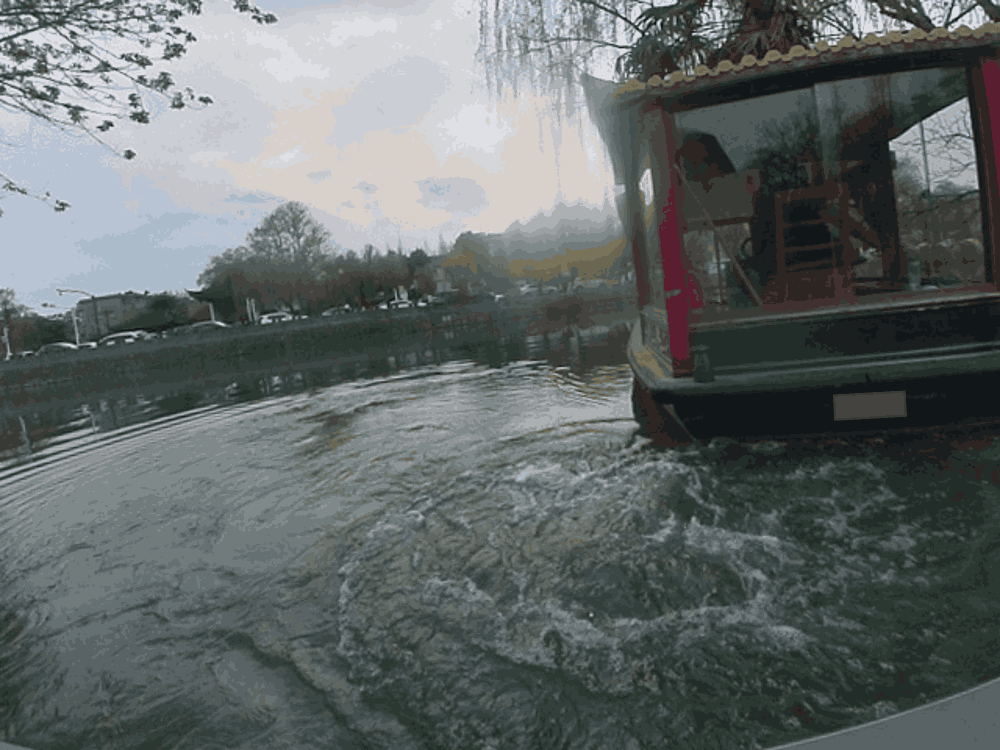}}
\caption{The example of ASVs motion vibrations caused by stern waves.}
\label{vib}
\vspace{-0.1cm}
\end{figure}
The data was collected by our ASV "Xi", equipped with cameras with a resolution of $480\times640$. The dataset was collected in varying weather and lighting conditions, including mirrors, dynamic surface textures, cluttered backgrounds, and motion vibrations interference, as demonstrated in Fig. \ref{datasetex} and Fig. \ref{vib}. 10 sequences that consist of 5530 frames, illustrating sunny daytime, midnight, dusk of both inland river and lake are involved. The dataset was clustered to training, validation, and testing by 6:2:2. We evaluated our model and traditional baselines and cutting-edge techniques for both image and video segmentation tasks on the proposed dataset.

\subsection{Training Settings}
Our experiments primarily use Nvidia 3060 GPUs equipped with 32GB of RAM. The implemented environment is set on PyTorch 1.8 and CUDA 11.1. For training, our model is trained for 1000 interactions. The initial learning rate was 0.0001, with the batch size of 4, optimized by the Stochastic Gradient Descent method with 0.9 momentum and 0.0001 weight decay. On the other hand, all the compared baseline models are trained and tested under recommended parameter settings, training interactions, and pre-trained backbones. 
During the experiments, we applied a random pick strategy to obtain pre-fusion frames. Specifically, we randomly selected two frames from four continuous former frames. This strategy was designed based on experiments that randomly and fixedly picked frames ranging from 32 to 2 previous frames. The randomly picking 2 from the 4 previous frames strategies worked out better.

\subsection{Evaluation Procedures}
 In previous works, the results of free space segmentation are tested on the whole scope. However, measuring the performance of free space segmentation models over the entire image scope is not suitable for the ASVs free space segmentation tasks. The size of the background areas is more extensive than in the lane zones. Therefore, a change in nearshore prediction would have little impact on the overall evaluation output. To improve this, we introduce a strategy to calculate the indicators in selected zones where the nearshore zones are enlarged, resulting in a more proper evaluation. The target zone is determined by cutting out the areas below the shoreline contours, as shown in Fig. \ref{compare}. This way, the unrelated background zones are ignored while the target free space area is emphasized.
 
\begin{figure}[!ht]
\vspace{-0.1cm}
\centering
\includegraphics[width=3.4in]{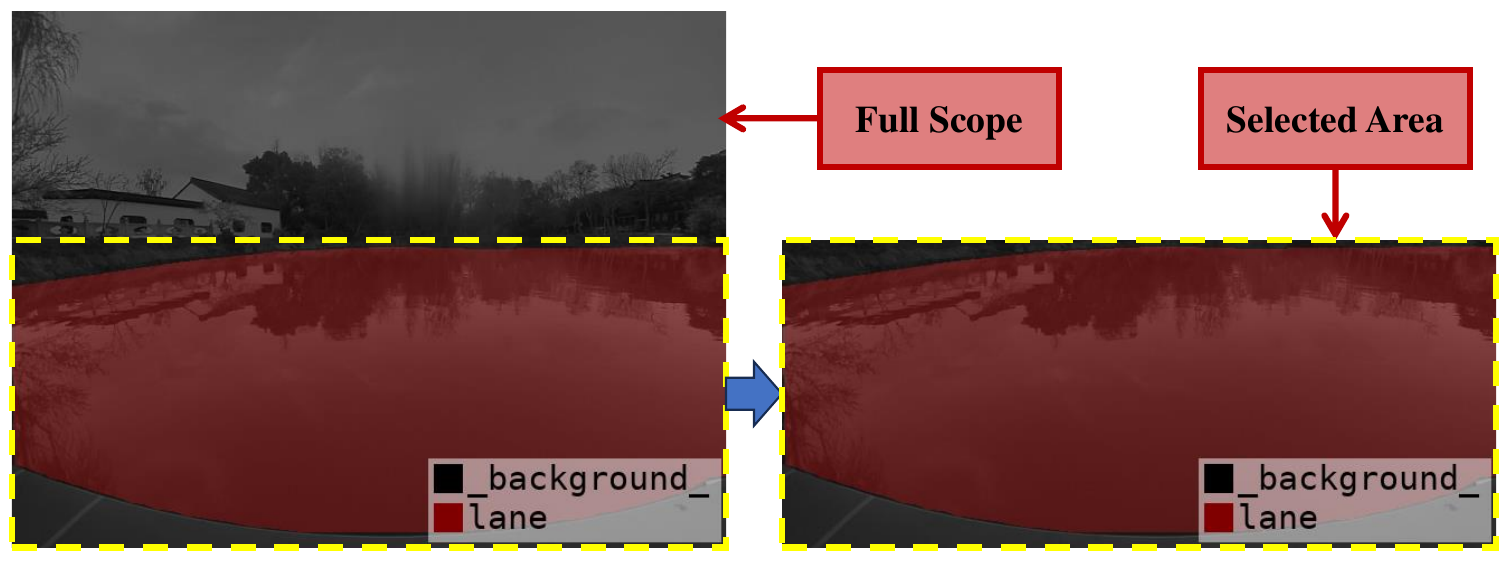}
\caption{The evaluation procedure that select the target free space zone from the whole scope.}
\label{compare}
\vspace{-0.1cm}
\end{figure}
\subsection{Baseline Comparison}
To evaluate the performance of our model, we compared it with image semantic segmentation baseline models, including baseline model Deeplabv3Plus \cite{chen2018encoderdecoder}, and state-of-the-art models Segformer\cite{xie2021segformer} and SETR-L \cite{Zheng_2021_CVPR}. In addition, we consider the general large visual model, making a comparison with the large model Segment Anything\cite{kirillov2023segment} released by Meta. As for video segmentation, we made the comparison to the baseline video segmentation model MasktrackRCNN\cite{yang2019video}, CrossVis\cite{steed2020crossvis}, and ETC \cite{liu2020efficient}.
\begin{figure}[htbp]
\centering
\includegraphics[width=3.5in]{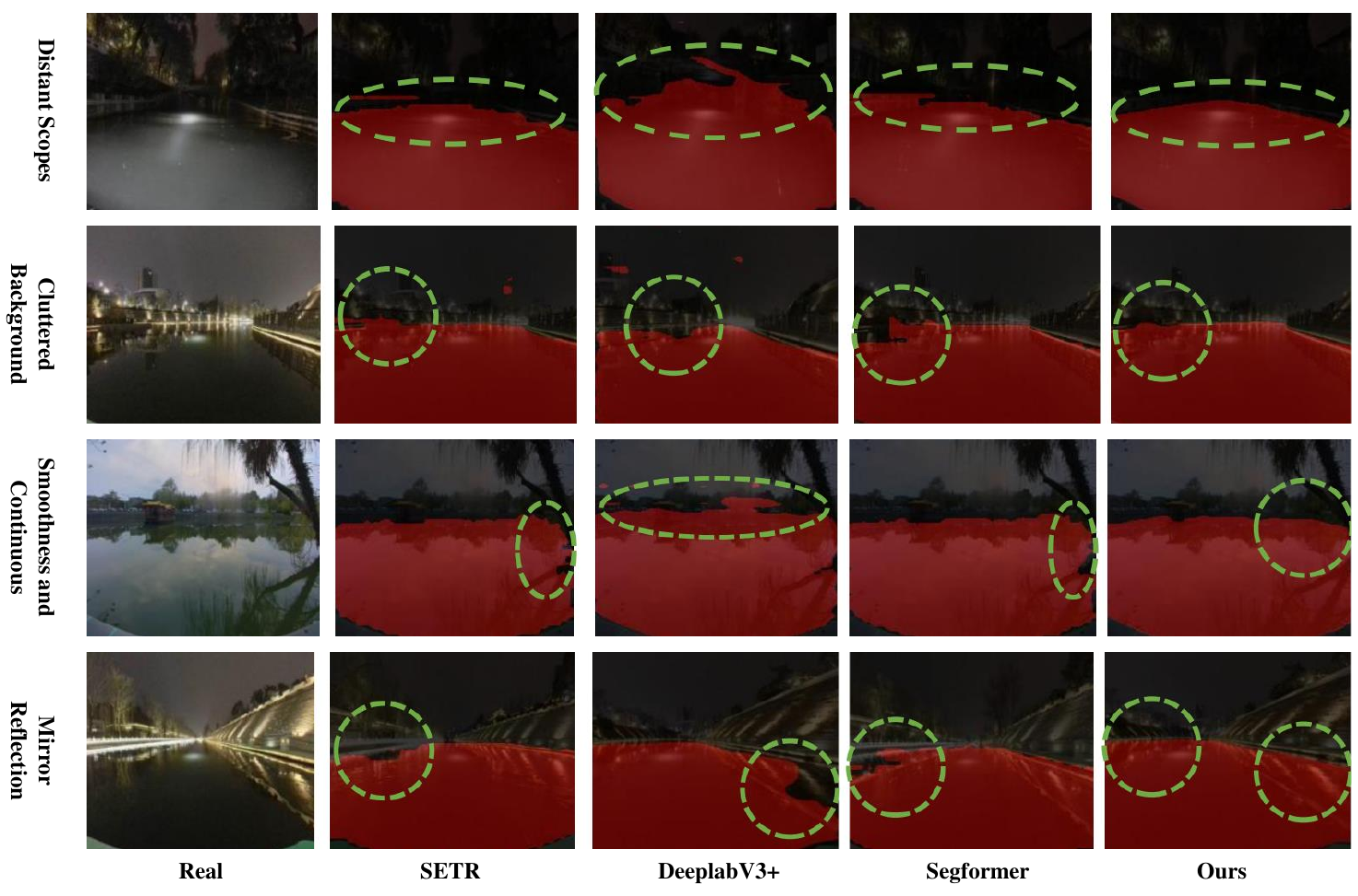}
\caption{The visualization comparison between the baseline and our model, our work shows better results.}
\label{fig_base}
\vspace{-0.1cm}
\end{figure}
The comparison results with the baseline models are shown in Table \ref{table1.1}.
\begin{table}[!htbp]
\center
\caption{Comparison with Semantic Segmentation Baselines\label{table1.1}}
\resizebox{1.0\columnwidth}{!}{
\begin{tabular}{lllllll}
\hline
Type&Model&MioU/selected&MioU/all
\\
\hline
 \multirow{4}{*}{Image}
     &Segformer&92.48&97.78
    \\
     &SETR&84.92&94.84
    \\
     &DeeplabV3+&86.265&95.40
    \\
     &SAM&71.16&84.93
\\
\hline
\multirow{3}{*}{Video}
     &ETC&88.62&92.11
    \\
     &MasktrackRCNN&80.01&86.63
    \\
     &CrossVis&90.01&92.12
    \\
\hline
&\textbf{Ours}&\textbf{94.13}&\textbf{98.11}
\\
\hline
\end{tabular}} 
\end{table}

Our approach can achieve better performance in ASVs free space segmentation tasks with appropriate real-time inference speed. Compared to image segmentation models, our work shows higher MioU than existing baselines in both whole and selected scopes. As for the comparison with video segmentation models, ours can achieve better outputs with relatively good computational efficiency. Regarding visualized results, as shown in Fig. \ref{fig_base}, our model demonstrates good performance in mirror reflection interference, cluttered background, and distant scenes. While others show zigzag outputs along the shoreline zones, our proposed model can generate free space segmentation results with smooth edges.  

According to the results, transformer-based models, such as SETR and Segformer, can obtain scores relatively well on evaluation indicators but struggle with complex scenes and produce zigzag shoreline segments in some cases. On the other hand, although DeeplabV3+ performed well in most common segmentation tasks, it lacks the capability of coping with surface challenges such as mirror reflections. In contrast, our model effectively deals with the aforementioned issues and outputs better scores. To further demonstrate their capability in dealing with mirror images, reflections and motion vibrations, we have generated an attention heat map\cite{selvaraju2017grad} in Fig. \ref{fig_heatmap}. The results clearly showcase how our work solves the segmentation challenges in nearshore areas. The attention in these zones is enhanced, resulting in accurate segmentation of reflections and instance lighting.
%
\begin{figure}[htbp]
\vspace{-0.1cm}
\centering
\includegraphics[width=3in]{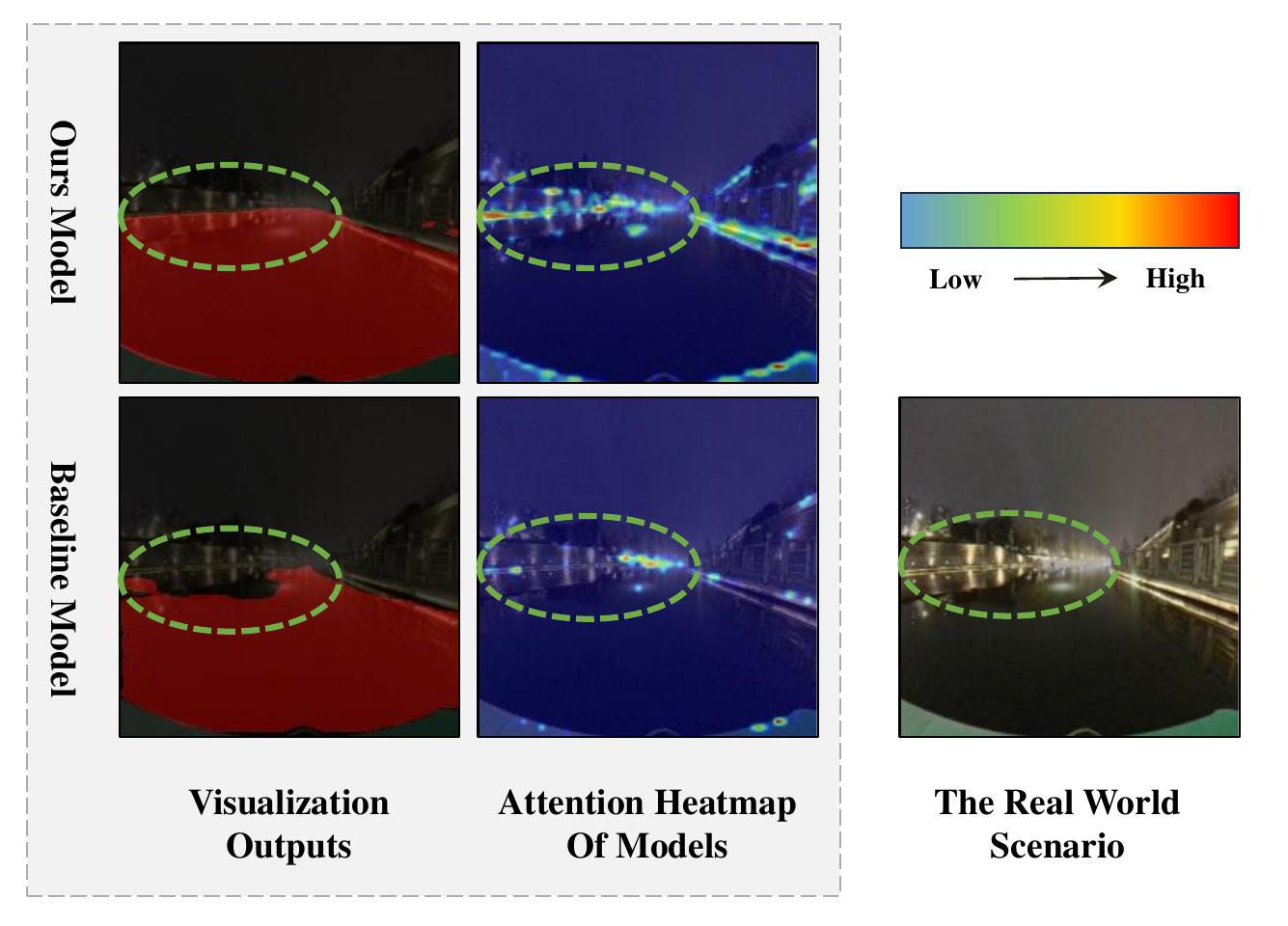}
\caption{The visualization comparison between the best-performed baseline and our model, our work shows high-accuracy.}
\label{fig_heatmap}
\vspace{-0.1cm}
\end{figure}
\subsection{Ablation Test}
To evaluate the performance of each module, we test the model without the temporal position encoder (described as $\mathrm{TPE}$ in the table), without the multi-head cross attention (described as $\mathrm{MAN}$ in the table), without the deformable convolution (described as $\mathrm{DCN}$ in the table), and without the Contour Loss (described as $L_{con}$ in the table). Our approaches show reliable performance on the modules above, the detailed results of ablation are as follows Table \ref{table2.1}.

\begin{table}[htbp]
\center
\caption{The results of module ablation experiments\label{table2.1}}
\resizebox{1.0\columnwidth}{!}{
\begin{tabular}{lllll}
\hline
Model&MioU/selected&MioU/all&GFlops&Parameters/M\\
\hline
Without TPE&92.75&97.68&21.24&71.08
\\
Without MAN&92.27&97.58&21.00
&70.51
\\
Without DCN&86.25&80.52&20.48
&58.13
\\
Without $L_{con}$&92.10&95.01&21.24&73.59
\\
All&94.13&98.11&21.24&73.59
\\
\hline
\end{tabular}}
\end{table}

The ablation test results indicate the efficiency of enhancing the relationship between previous features and augmenting the interested features before and during the fusion procedure. Each module contributes well to the efficiency and accuracy of the free space segmentation tasks. The attention mechanisms lead to the evident decline of evaluation indicators. It is worth pointing out that the significant influence of the deformable convolution module shows that our work can sufficiently cope with the motion vibrations, and mitigate the noises brought by the temporal fusion mechanism.

\subsection{Robustness Test}
When conducting real-time missions with the ASVs, it is essential to consider dynamic interference. To test the system's robustness under dynamic interference, we conducted a test that accounted for frame drops and backward driving. The results of the dynamic interference test are presented in the following Table \ref{table2.2}.

\begin{table}[htbp]
\center
\caption{The results of robustness test under dynamic interference such as drops and reverse.\label{table2.2}}
\resizebox{1.0\columnwidth}{!}{
\begin{tabular}{lllll}
\hline
Sequence&Direction&Drops&MioU/selected&MioU/all\\
\hline
\multirow{4}{*}{Seq4(Night)}
    &Forward&None&92.59&98.09
\\
    &Backward&None&92.35&97.98
\\
    &Forward&1/7&92.54&98.09
\\
    &Backward&1/7&92.1&97.91
\\
\hline
\multirow{4}{*}{Seq5(Day)}
    &Forward&None&92.07&98.09
\\
    &Backward&None&92.03&97.65
\\
    &Forward&1/7&92.02&98.04
\\
    &Backward&1/7&91.93&97.51
\\
\hline
\end{tabular}}
\end{table}

The results of the robustness experiments show that our model can produce reliable segmentation results even when subjected to frame drops or backward movements during both daytime and nighttime conditions. The evaluation indicators exhibit minimal declines despite such noises. The results demonstrate the good robustness of our model.

\subsection{Real World Experiment}
To further evaluate the performance of the proposed model, we test the performance on real-world ASVs platforms and scenarios, as illustrated in Fig. \ref{fig_realworld}. The experimental ASVs platform is equipped with five cameras, utilizing Nvidia Jetson Orin NX as the computing platform. 

\begin{figure}[htbp]
\vspace{-0.1cm}
\centering
\includegraphics[width=3.25in]{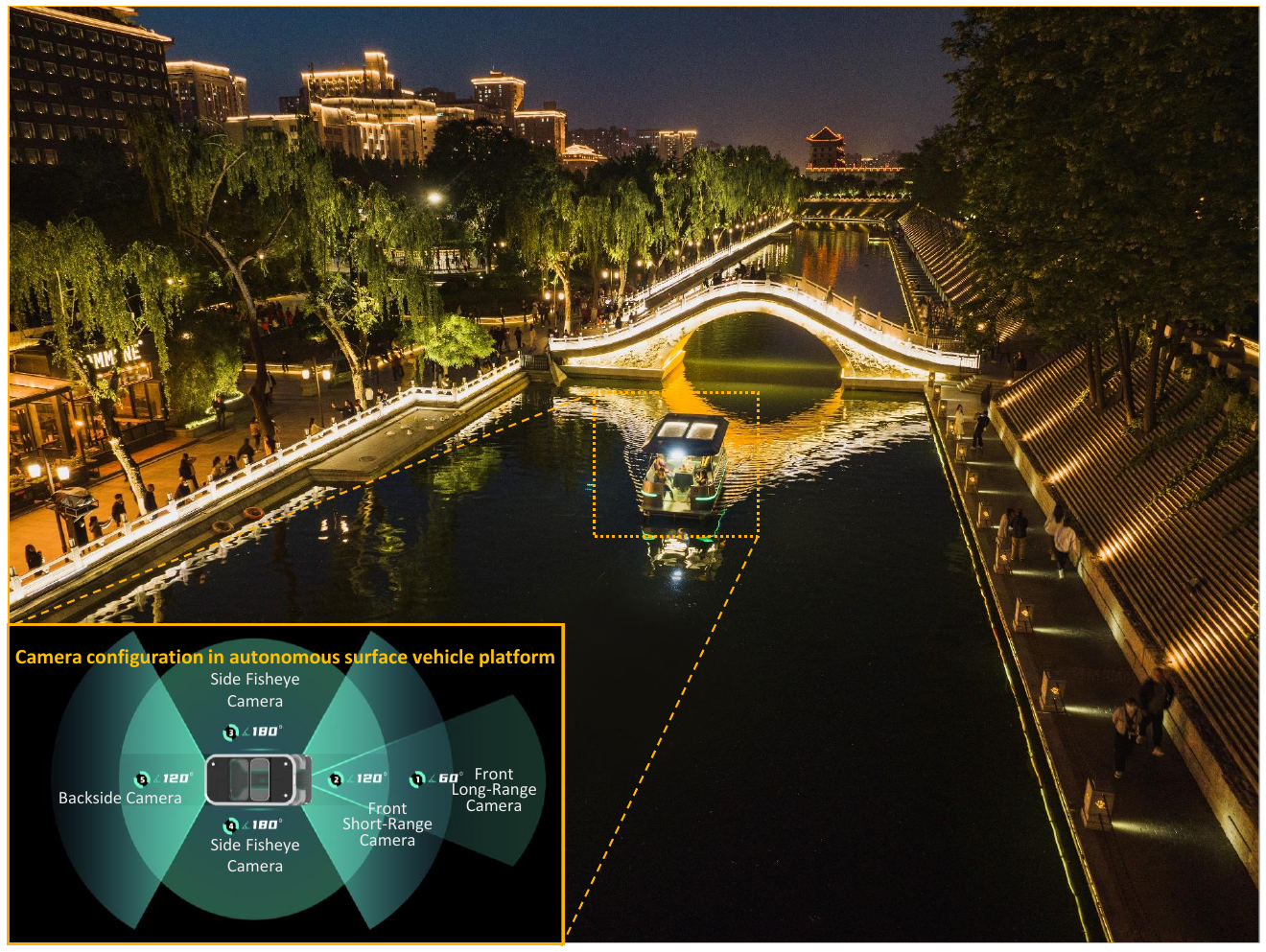}
\caption{The illustration of our real-world experiments, the applied platform was equipped with five cameras, including long and short range cameras, backside cameras, and two fisheye cameras.}
\label{fig_realworld}
\vspace{-0.3cm}
\end{figure}

With an average speed of 0.8m/s along the shoreline, our model is able to obtain robust and reliable online free space segmentation in real time, outputting results with five cameras' inputs, including long and short range cameras, backsides cameras, and two fisheye cameras, as illustrated in Fig. \ref{fig_5}. While inference the inputs of five cameras at the time, our model could achieve real-time inference of 4.72 FPS, and 23.6 FPS on a single camera. The test scenes were not learned by the model before, and the outputs indicate the robustness and efficiency of our work in real-world applications.
\begin{figure}[htbp]
\centering
\includegraphics[width=3.5in]{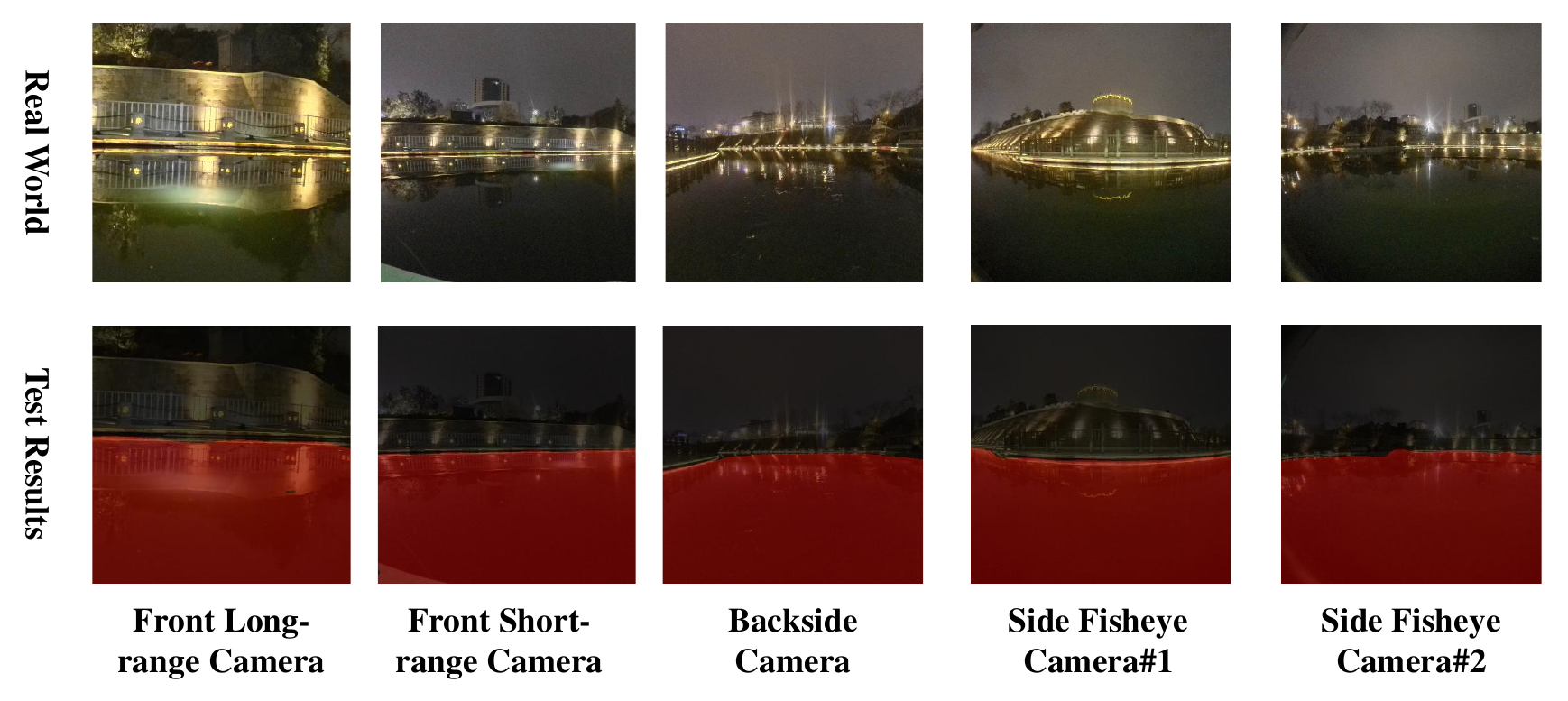}
\caption{The segmentation results of five cameras at the same time.}
\label{fig_5}
\vspace{-0.1cm}
\end{figure}
\section{Conclusion}
This work focuses on mirror reflection, dynamic surface textures, and ASVs motion vibration problems during the free space segmentation tasks. To obtain higher accuracy, we propose a visual temporal fusion based semantic segmentation model for free space segmentation tasks. With previous-to-now alignment and augmented fusion modules, our work can efficiently handle ASVs free space segmentation tasks under diverse conditions. Next, we introduce a new loss function for calculating loss based on contour distance to optimize the training process. The model is evaluated in selected scopes to mitigate the imperfection of the MioU test on full scope. We also introduce a video ASVs free space segmentation dataset to benefit the ASVs research community, and evaluate the proposed model on it. The results show the efficiency and robustness of our work.

In the future, we plan to make the work more applicable and accurate. On the one hand, the inference speed of the proposed model could be improved. Therefore, our next step will be to lighten our model with further experiments and analysis to achieve faster calculation. On the other hand, there is potential to expand the capability of our model. It is possible to make the presented work adapted to multitasks in the future. 

\bibliographystyle{IEEEtran}
\small\bibliography{ref}

\end{document}